%% file: main.tex
\documentclass[letterpaper, 10 pt, conference]{ieeeconf}  

\IEEEoverridecommandlockouts                              

\usepackage{graphicx}
\graphicspath{{figs/}}
\usepackage{amsmath}
\usepackage{amssymb}
\usepackage{algorithm}
\usepackage{algpseudocode}
\usepackage{times}
\usepackage{bm}
\usepackage{physics}
\usepackage{amsmath}
\newcommand{\figref}[1]{Fig.~\ref{figure:#1}}

\title{\LARGE \bf
A Morphing Aerial Robot With Thruster-Integrated \\ Flexible Continuum Links for Shape Adaptive Aerial Manipulation
}

\author{
        Eri Sawada$^{1}$, Kazuki Sugihara$^{1}$, Ayano Miyamichi$^{1}$, Kunio Kojima$^{1}$, and Kei Okada$^{1}$
\thanks{$^{1}$The authors are with the Department of Mechano-Infomatics, The University of Tokyo, Bunkyo-ku, Tokyo 113-8656, Japan. [e-sawada, sugihara, miyamichi, k-kojima, k-okada]@jsk.imi.i.u-tokyo.ac.jp}%
}

\begin{document}

\maketitle
\thispagestyle{empty}
\pagestyle{empty}

\begin{abstract}
\input src/abst.tex
\end{abstract}

\input src/intro.tex
\input src/related_works.tex
\input src/mech.tex

\input src/model.tex

\input src/control.tex
\input src/experiment.tex
\input src/conclusion.tex

\addtolength{\textheight}{-12cm}   


\bibliographystyle{unsrt}
\bibliography{main}

\end{document}

%% file: src/abst.tex
In recent years, aerial manipulation has attracted increasing attention as a key to expand the application of aerial robots. In this work, we focus on two major research directions for achieving versatile aerial manipulation: (i) acquiring high environmental adaptability using soft manipulators, and (ii) expanding the feasible wrench space by distributing thrusters along the manipulator. However, no aerial robot has simultaneously satisfied these two requirements. Therefore, in this paper, we propose a morphing rotor-distributed aerial robot with flexible continuum links that achieves both high shape adaptability and an expanded wrench space. The flexible continuum links function as soft manipulators, passively conforming to the shape of the environment, while the thrusters distributed along the continuum links expand the feasible thrust wrench space and enable the end-effector to exert large interaction forces. To realize the proposed robot, it is essential to suppress vibrations of the lightweight continuum links. Thus, we develop a composite leaf-spring structure that provides both high torsional and vertical stiffness, and vibration-suppressing control methods. Using these implementations, we demonstrate stable flight and a variety of aerial manipulation tasks. To the best of our knowledge, this is the first work to realize aerial manipulations using flexible links with an integrated thruster.

%% file: src/intro.tex
\section{INTRODUCTION}

\begin{figure}[!t]
  \centering
  \includegraphics[width=\linewidth]{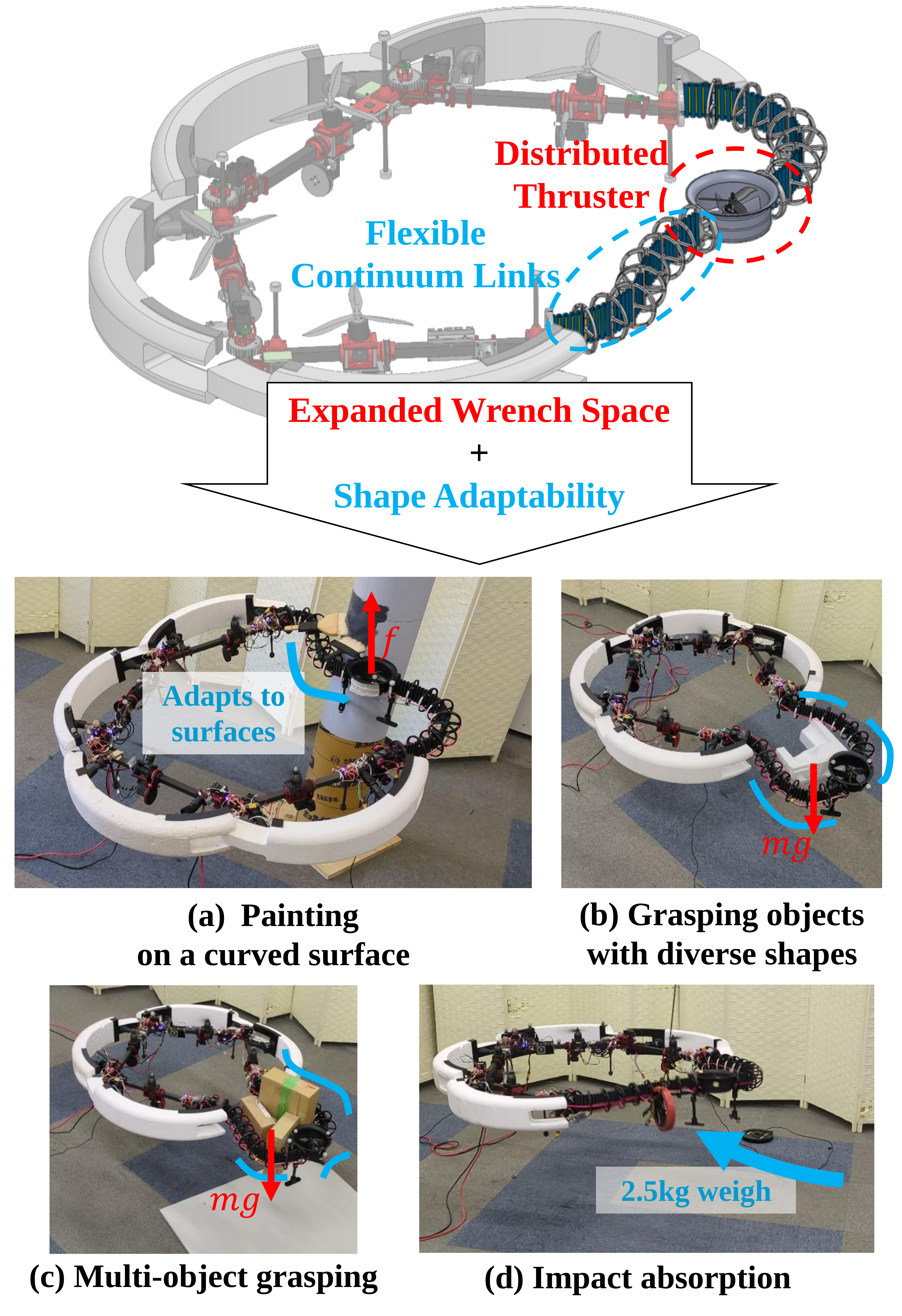}
  \caption{The proposed morphing aerial robot features flexible continuum links with an integrated thruster. This combination provides both high adaptability and an expanded wrench space. Leveraging these properties, the robot performed aerial manipulation tasks such as (a) painting on a curved surface, (b) grasping objects of various shapes, and (c) multi-object grasping, and also demonstrated (d) impact absorption.}
  \label{figure:overview}
  \vspace{-10pt}
\end{figure}

For many years, aerial robots have been used mainly for non-contact tasks such as surveillance and remote sensing. In recent years, however, aerial manipulation---performing tasks while making contact with the environment---has advanced, and applications such as transportation and contact-based inspection in higher places or hazardous areas are increasingly expected \cite{Ruggiero2018aerialmanipulation}. In this work, we focus on two major approaches that have been explored to realize versatile aerial manipulation. The first approach is achieving high environmental adaptability using soft manipulators \cite{Ubellacker2024softdrone, ZhichaoLiu2022softgripper, Wang2025spirobs, ruiz2022sophie, broers2022design}. High adaptability allows a manipulator to passively conform to various geometries and enables aerial manipulation that can robustly adapt to complex and diverse real-world environments. The second approach is to distribute rotors along the manipulator to achieve an expanded thrust wrench space \cite{zhao2022forceful, zhao2017whole}. Here, the thrust wrench space denotes the set of achievable forces and torques about the center of gravity (CoG) while providing the thrust required for gravity compensation. An expanded wrench space enables the stable execution of tasks that require large interaction forces with the environment, such as painting, window cleaning, and grasping heavy objects.

To leverage both advantages simultaneously, we propose a new rotor-distributed morphing aerial robot with flexible continuum links, as shown in \figref{overview}. The flexible continuum links function as a manipulator, providing high shape adaptability for aerial manipulation. Moreover, a thruster mounted on the flexible links contributes to an expanded wrench space and enable the end-effector to exert large interaction forces.

However, lightweight flexible continuum links in aerial robots can easily deform in unintended directions and are prone to vibration during flight. To overcome these issues, we propose a composite leaf-spring structure for the flexible links and two vibration-suppression control strategies. The proposed composite leaf-spring structure combines flat and folded (U-shaped) PLA leaf springs to achieve both high torsional stiffness and vertical stiffness while maintaining horizontal flexibility. The proposed control strategies consist of (i) thrust-difference minimization, which suppresses variations in commanded thrust, and (ii) low-rate updates of the wrench-allocation matrix, which filter high-frequency variations in the estimated robot configuration.

Using these design and control strategies, we build the proposed aerial robot and achieve stable flight. By leveraging its flexibility and an expanded wrench space, we demonstrate a variety of aerial manipulation tasks, such as painting on curved surfaces (\figref{overview}(a)), grasping objects with diverse shapes (\figref{overview}(b)), and multi-object grasping (\figref{overview}(c)).

The main contributions of this work are as follows:
\begin{itemize}
    \item We propose a morphing aerial robot with flexible continuum links and distributed thrusters, enabling both high shape adaptability and an expanded thrust wrench space.
    \item We design a lightweight composite leaf-spring structure for the flexible continuum links that achieves both torsional and vertical stiffness while maintaining horizontal flexibility.
    \item We develop a vibration-suppressing control scheme that attenuates high-frequency components in thrust allocation and mitigates vibrations of the flexible links.
    \item We demonstrate the effectiveness of the proposed method through experiments including painting on a curved surface, grasping objects of various shapes, and multi-object grasping.
\end{itemize}

The rest of this paper is organized as follows. Section II describes related works. Section III describes the design of the proposed flexible continuum links. Section IV describes the modeling of the morphing aerial robot with flexible continuum links. Section V describes the control method to suppress the vibration of the flexible links. Section VI describes the experimental results, and Section VII concludes the paper.

%% file: src/related_works.tex
\section{RELATED WORKS}
This work aims to present an aerial robot that simultaneously achieves both high shape adaptability and an expanded wrench space. In this section, we review prior studies from these two perspectives.

\subsection{Robots With High Environmental Adaptability}
In recent years, soft robots have gained significant attention for their high compliance and shape adaptability. For example, soft grippers and continuum robots inspired by biological trunks and tentacles are well suited for navigating through narrow and cluttered environments and for grasping objects with diverse shapes, owing to their intrinsic flexibility and shape conformity \cite{walker2013continuous, shintake2018soft}.

These soft manipulators have also been adopted in aerial robots and have been shown to expand the applicability and robustness of aerial manipulation. For instance, mounting a soft gripper on an aerial robot enables aerial grasping while in motion \cite{Ubellacker2024softdrone, ZhichaoLiu2022softgripper}. Moreover, several studies using soft arms have demonstrated enhanced perching capabilities, in which an aerial robot attaches to and remains on a structure such as a wall or branch to rest or perform a task. Thanks to their compliance and shape adaptability, soft arms enable perching on targets with a wide range of geometries \cite{ruiz2022sophie, broers2022design}. Building on these studies, our work integrates distributed thrusters into a soft manipulator. This integration preserves shape adaptability while expanding the wrench space. As a result, the robot can generate larger interaction forces and perform tasks such as painting on curved surfaces and grasping objects with diverse geometries.

\subsection{Aerial Robot With an Expanded Wrench Space}
A straightforward configuration for aerial manipulation is to mount a manipulator on an underactuated quadrotor without a thrust-vectoring mechanism \cite{fumagalli2014developing, hunt20143d}. While this configuration is mechanically simple, it is difficult to generate interaction forces other than upward thrust, such as large moments or horizontal forces.

To address this issue, some studies expanded the wrench space by increasing the number of thrusters and optimizing their placement\cite{park2018odar}. Building on this direction, fully actuated aerial robots with thrust-vectoring mechanisms have also been proposed. By allowing the thrust directions to rotate, these platforms increase the number of control inputs without increasing the number of thrusters, thereby enabling wrench generation in horizontal directions and expanding the wrench space \cite{tognon2019truly, papachristos2014efficient}. However, when the manipulator is extended or grasps a heavy object, the overall CoG may move outside the horizontal-plane projection of the region spanned by the thrusters. In such cases, maintaining stable flight often becomes difficult even with thrust-vectoring mechanisms.

More recently, a different approach has been proposed: distributing thrusters along the manipulator \cite{zhao2022forceful, zhao2017whole}. Rather than separating the flight platform and the manipulator, these systems perform manipulation by morphing the body itself. Under this strategy, the system can be viewed as a manipulator equipped with distributed thrusters. Thanks to the thrusters mounted along the manipulator, the CoG can remain surrounded by rotors even when the arm is extended or carrying a heavy payload. This expanded wrench space has enabled the stable execution of tasks that require large interaction forces, such as valve turning \cite{zhao2022forceful} and grasping heavy objects \cite{zhao2017whole}.

%% file: src/mech.tex
\section{DESIGN OF FLEXIBLE CONTINUUM LINKS}

In the proposed aerial robot, the flexible continuum link must satisfy the following three requirements:

\begin{itemize}
  \item lightweight design for flight,
  \item high vertical stiffness to prevent vibration caused by thrust, while remaining flexible in the horizontal direction to allow two-dimensional deformation,
  \item a structural restoring force in the horizontal direction to stabilize the shape of the underactuated continuum links.
\end{itemize}

\begin{figure}[t]
  \centering
  \includegraphics[width=\columnwidth]{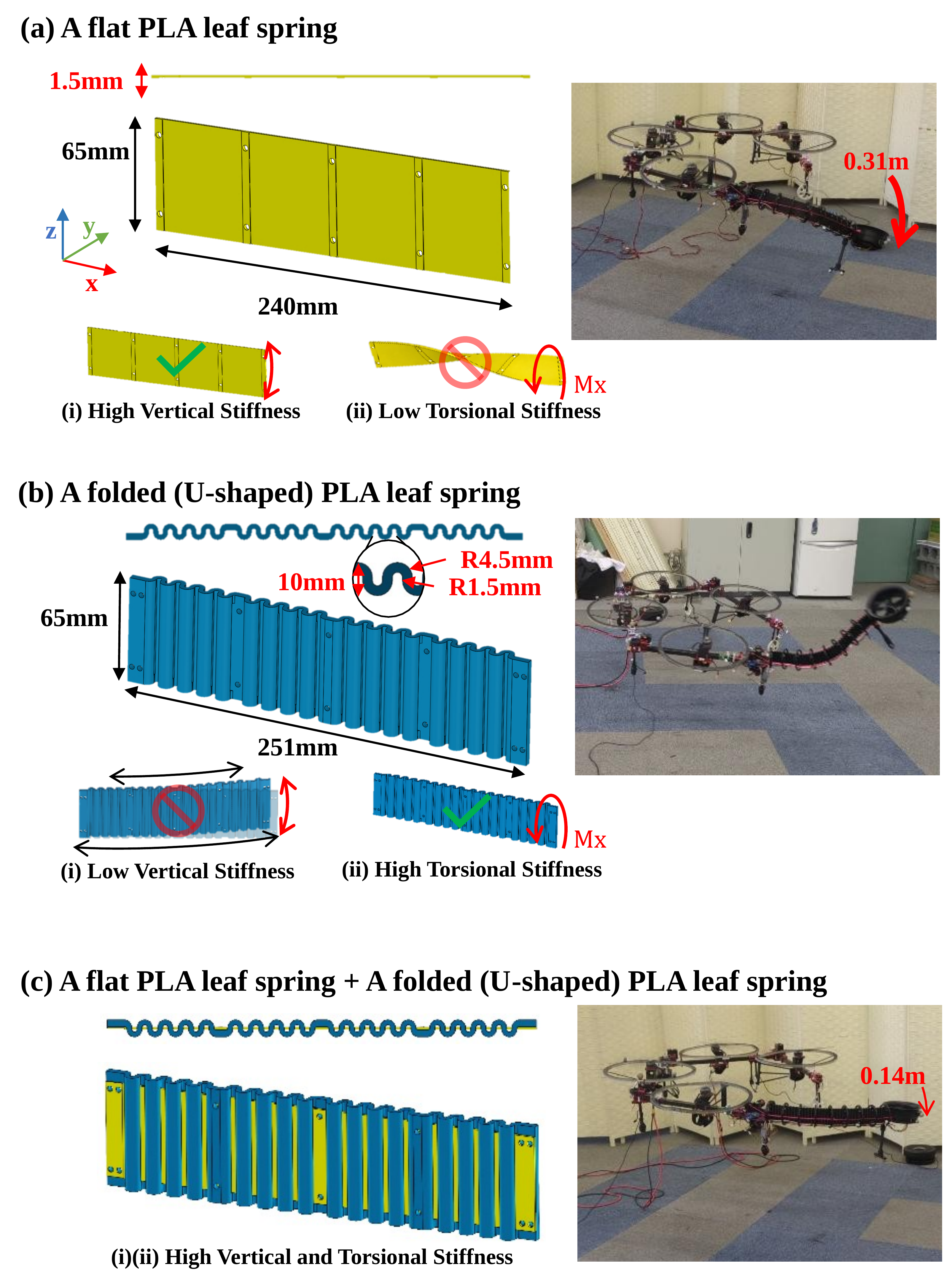}
  \caption{Comparison of flexible continuum-link designs: (a) With a flat PLA leaf spring, the structure twists and sags during flight. (b) With a folded (U-shaped) PLA leaf spring, torsion is suppressed; however, the vertical stiffness is insufficient, causing large oscillations immediately after takeoff. (c) With the proposed composite leaf spring, both torsional and vertical stiffness are achieved, enabling stable in-air morphing.}
  \label{figure:backbone_comparison}
\end{figure}

A simple design that satisfies these requirements is a flat leaf spring, as shown in \figref{backbone_comparison}(a). The flat PLA leaf spring, fabricated using a 3D printer, is lightweight and provides high stiffness against bending in the vertical \(z\)-direction while remaining compliant in the intended morphing direction, i.e., the \(y\)-direction defined in \figref{backbone_comparison}. It also provides a structural restoring force for shape stabilization. However, the flat leaf spring is prone to torsional deformation about the link axis (the \(x\)-axis). During flight, the continuum link tends to twist and sag, causing the rotor thrust axis to deviate from the vertical direction (see the flight snapshot in \figref{backbone_comparison}(a)). This results in a mismatch between the actual robot configuration and the intended model, degrading flight stability and preventing the robot from realizing the intended wrench space.

To suppress torsion about the \(x\)-axis, we developed a hollow folded (U-shaped) structure, as shown in \figref{backbone_comparison}(b). The increased sectional thickness improves torsional stiffness about the \(x\)-axis while preserving compliance in the morphing \(y\)-direction. However, because the upper and lower parts can extend and contract, the structure remains compliant in the vertical \(z\)-direction, leading to large oscillations immediately after takeoff.

Based on these observations, we developed the composite leaf spring shown in \figref{backbone_comparison}(c), which achieves both high torsional stiffness about the \(x\)-axis and high stiffness against bending in the vertical \(z\)-direction. The structure combines a flat leaf spring and a folded leaf spring by introducing a slit along the center of the folded spring and inserting the flat spring into the slit. As a result, the link retains compliance in the intended morphing direction while suppressing undesired torsion and vertical bending. This enables the robot to maintain the intended configuration and realize the desired wrench space.

%% file: src/model.tex
\section{MODELING OF THE MORPHING AERIAL ROBOT WITH FLEXIBLE CONTINUUM LINKS}
\begin{figure}[t]
  \centering
  \includegraphics[width=\columnwidth]{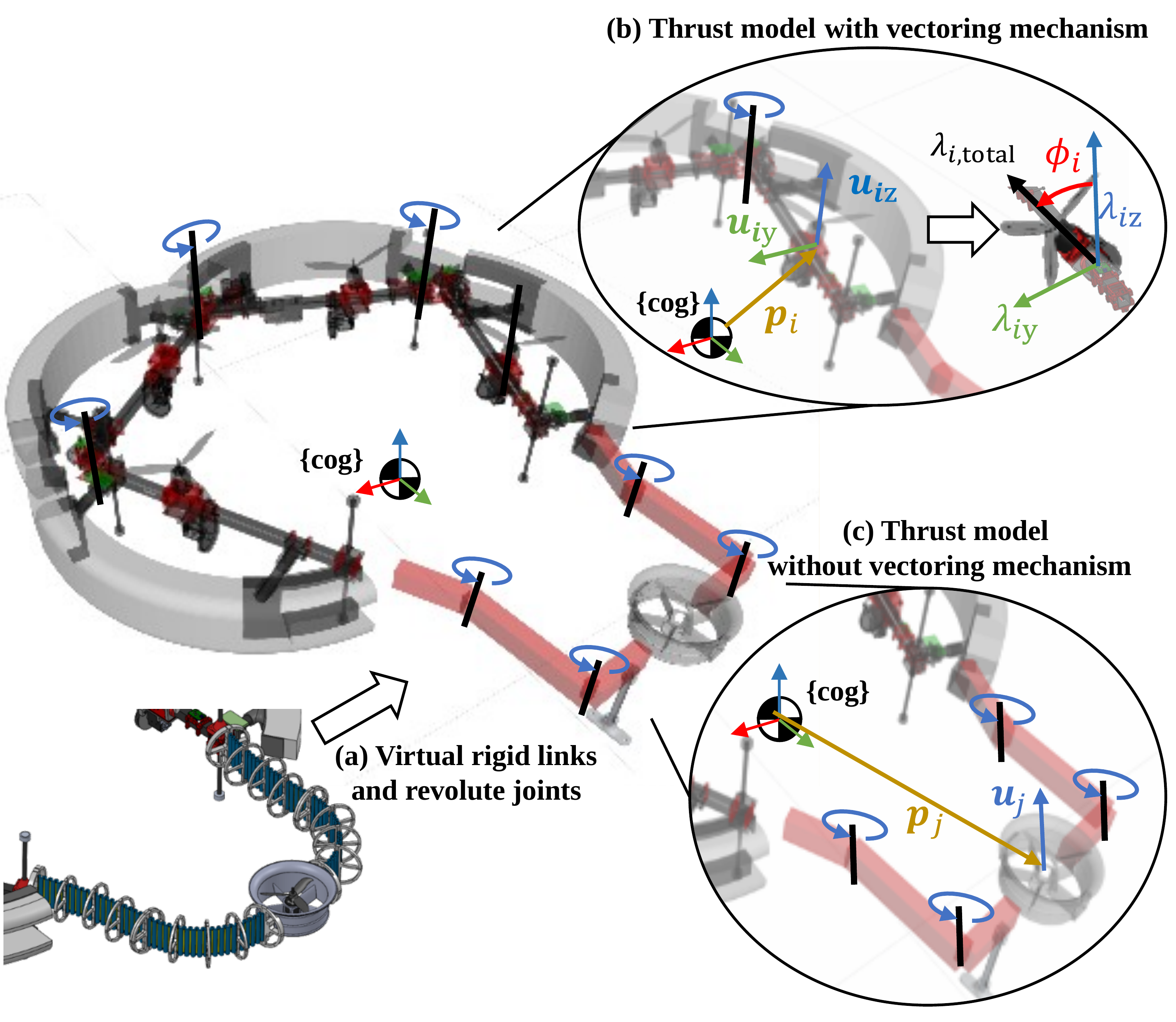}
  \caption{Kinematic model of the robot with flexible continuum links and thrust models with and without the thrust-vectoring mechanism. (a) The flexible continuum link is approximated as a chain of rigid links connected by multiple virtual rotational joints. (b), (c) In modeling the thrust-vectoring mechanism, we introduce virtual thrusters that produce vertical and horizontal thrust components independently, enabling a linear representation of the thrust model.}
  \label{figure:thrust_model}
\end{figure}
This section describes the kinematic and thrust models of the proposed morphing aerial robot equipped with flexible continuum links.

Several modeling approaches have been proposed for continuum structures, such as the Piecewise Constant Curvature (PCC) model \cite{webster2010design}. In this work, however, to ensure real-time computation on an onboard computer with limited resources, we approximate the flexible link as a chain of rigid links connected by multiple virtual rotational joints, as shown in \figref{thrust_model}(a). The robot model constructed here is used to compute the CoG, the position of each thruster $\bm{p}_i\in\mathbb{R}^3$, and its thrust-axis unit vector $\bm{u}_i\in\mathbb{R}^3$, expressed in the CoG frame. The virtual joint angles are determined from commanded deformation inputs or loop-link kinematic constraints that enforce consistency in the position and orientation at both ends of the links.

The robot is equipped with $N_{\rm {r}}$ thrusters with a thrust-vectoring mechanism, as shown in \figref{thrust_model}(b), and $N_{\rm {s}}$ thrusters without a thrust-vectoring mechanism, as shown in \figref{thrust_model}(c). While the thrust-vectoring mechanism increases the control degrees of freedom, applying it to thrusters mounted on the flexible continuum links can induce vibration. For this reason, the thrusters attached to the flexible continuum links do not have a thrust-vectoring mechanism.

Because a thrust-vectoring mechanism changes the rotor-axis direction nonlinearly, the mapping from control inputs to the generated wrench about the CoG becomes nonlinear. To obtain a linear approximation, we model each vectorable thruster as two virtual thrusters that independently generate a vertical component $\lambda_{iz}$ and a horizontal component $\lambda_{iy}$, as shown in \figref{thrust_model}(b), and define $\bm{u}_{iz}$ and $\bm{u}_{iy}$ as the rotor-axis unit vectors of the vertical and horizontal virtual thrusters, respectively. The resultant thrust magnitude $\lambda_{i,\rm total}$ and the vectoring angle $\phi_i$ are calculated as follows.
\begin{gather}
    \lambda_{i, \rm{total}} = \sqrt{\lambda_{iz}^2 + \lambda_{iy}^2}, \\
    \phi_i = \mathrm{atan2}(\lambda_{iy}, \lambda_{iz}).
\end{gather}
Let $\bm{\lambda}_{\rm full}\in\mathbb{R}^{N}$ denote the thrust vector including the virtual thrust components introduced by the vectoring mechanisms, where $N=2N_{\rm r}+N_{\rm s}$. \(\bm{\lambda}_{\mathrm{full}}\) is defined as
\begin{gather}
\small
\bm{\lambda}_{\rm full} =
\begin{bmatrix}
    \underbrace{
        \lambda_{1z} \; \lambda_{1y} \; \cdots \; \lambda_{N_r z} \; \lambda_{N_r y}
    }_{2N_r}
    \;
    \underbrace{
        \lambda_{N_r+1} \; \cdots \; \lambda_{N_r+N_s}
    }_{N_s}
\end{bmatrix}^{\rm{T}}
\normalsize
\end{gather}
For a given joint configuration \(\bm{q}\), the total force \(\bm{f}_{\lambda}\) and total torque \(\bm{\tau}_{\lambda}\) about the CoG generated by \(\bm{\lambda}_{\mathrm{full}}\) are written as
\begin{gather}
    \begin{bmatrix}
        \boldsymbol{f}_{\rm{\lambda}} \\
        \boldsymbol{\tau}_{\rm{\lambda}}
    \end{bmatrix} 
    = Q(\bm{q}) \boldsymbol{\lambda}_{\rm{full}}.
\end{gather}
To define \(Q(\bm{q})\), we first introduce the wrench generated by a unit thrust applied at position \(\bm{p}\), along direction \(\bm{u}\), and \(\sigma\). \(\sigma\) represents the reaction torque coefficient with the sign representing the direction of rotation:
\begin{gather}
\bm{w}(\bm{u}, \bm{p}, \sigma)
=
\begin{bmatrix}
\bm{u} \\
\bm{p} \times \bm{u} + \sigma \bm{u}
\end{bmatrix}
\in \mathbb{R}^{6}.
\end{gather}
Using this notation, the wrench-allocation matrix is given by

\begin{gather}
Q(\bm{q}) =
\left[
\begin{array}{c}
    \bm{w}(\bm{u}_{1z}, \bm{p}_1, \sigma_1)^{\rm{T}} \\
    \bm{w}(\bm{u}_{1y}, \bm{p}_1, \sigma_1)^{\rm{T}} \\
    \vdots \\
    \bm{w}(\bm{u}_{N_rz}, \bm{p}_{N_r}, \sigma_{N_r})^{\rm{T}} \\
    \bm{w}(\bm{u}_{N_ry}, \bm{p}_{N_r}, \sigma_{N_r})^{\rm{T}} \\
    \bm{w}(\bm{u}_{N_r+1}, \bm{p}_{N_r+1}, \sigma_{N_r+1})^{\rm{T}} \\
    \vdots \\
    \bm{w}(\bm{u}_{N_r+N_s}, \bm{p}_{N_r+N_s}, \sigma_{N_r+N_s})^{\rm{T}}
\end{array}
\right]^{\!\rm{T}}
\begin{array}{l}
    \left.\vphantom{
    \begin{array}{c}
        \bm{w}(\bm{u}_{1z}, \bm{p}_1, \sigma_1) \\
        \bm{w}(\bm{u}_{1y}, \bm{p}_1, \sigma_1) \\
        \vdots \\
        \bm{w}(\bm{u}_{N_rz}, \bm{p}_{N_r}, \sigma_{N_r}) \\
        \bm{w}(\bm{u}_{N_ry}, \bm{p}_{N_r}, \sigma_{N_r})
    \end{array}
    }\right\}
    \, 2N_r
    \\[3.5em]
    \left.\vphantom{
    \begin{array}{c}
        \bm{w}(\bm{u}_{N_r+1}, \bm{p}_{N_r+1}, \sigma_{N_r+1}) \\
        \vdots \\
        \bm{w}(\bm{u}_{N_r+N_s}, \bm{p}_{N_r+N_s}, \sigma_{N_r+N_s})
    \end{array}
    }\right\}
    \, N_s
\end{array}.
\end{gather}
The first \(2N_r\) columns correspond to the virtual thrust components of the vectoring thrusters, and the remaining \(N_s\) columns correspond to the fixed-direction thrusters.

Using the computed net force and torque, the equations of motion about the CoG are given below. In this work, joint angles are position-controlled, and the robot is treated as a single rigid body for thrust allocation at each time step.
\begin{gather}
m\ddot{\bm{r}} = \bm{f}_{\rm{\lambda}} - m\bm{g}, \\ 
I\dot{\bm{\omega}} = - \bm{\omega} \times I\bm{\omega} + \bm{\tau}_{\rm{\lambda}}.
\end{gather}
where $\bm{r}$ is the CoG position in the world frame, $\bm{g}$ is the gravity vector, and $\bm{\omega}$ is the angular velocity. $\ddot{\bm{r}}$ and $\dot{\bm{\omega}}$ are the desired linear and angular accelerations computed by PID controllers. The desired linear acceleration $\ddot{\bm{r}}$ is calculated from the error between the desired center-of-gravity position and the model-computed center-of-gravity position in the world frame, while the desired angular acceleration $\dot{\bm{\omega}}$ is calculated from the attitude error of the center-of-gravity frame, which corresponds to that of the base-link frame.

%% file: src/control.tex
\section{VIBRATION-SUPPRESSING CONTROL}
\label{control}
\begin{figure}[t]
  \centering
  \includegraphics[width=\linewidth]{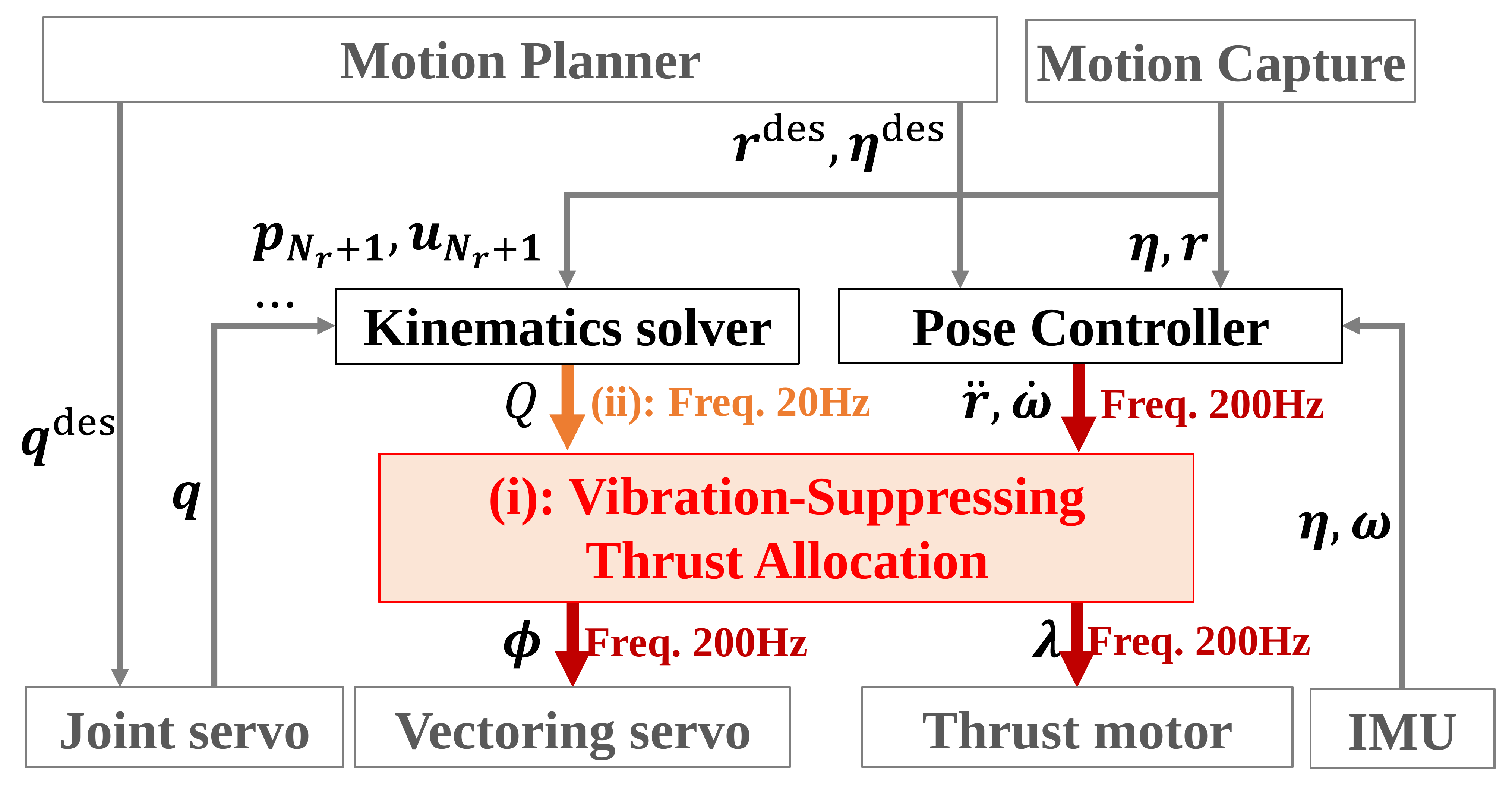}
  \caption{Architecture of the proposed vibration-suppressing control. (i) Vibration attenuation by thrust-difference minimization (Sec.~\ref{sec:thrust_diff_min}). (ii) Low-frequency update of the wrench-allocation matrix (Sec.~\ref{sec:reduce_update_hz_of_q}), in which the desired accelerations are computed by a high-frequency pose controller, while the wrench-allocation matrix \(Q\) is updated at a lower frequency.}
  \label{figure:control_system}
\end{figure}
In this section, we describe two control methods for suppressing vibrations of the flexible links. The control framework is shown in \figref{control_system}. 
The proposed control methods are evaluated through experiments in Sec.~\ref{sec:experiment_vibration_suppression}.

\subsection{Thrust Allocation With Thrust-Difference Minimization}
\label{sec:thrust_diff_min}
Next, we describe a thrust-allocation method for suppressing vibrations of the flexible links. The conventional approach computes the thrust command $\bm{\lambda}_{\rm full}$ by solving a linearly constrained quadratic programming that minimizes the thrust magnitude:
\begin{align}
    & \underset{\bm{\lambda}_{\rm{full}}}{\text{minimize}} 
    & & 
        \bm{\lambda}_{\rm{full}}^{\rm{T}}
        W
        \bm{\lambda}_{\rm{full}}\\
    & \text{subject to} 
    \label{eq:lambda_opt}
    && 
    \begin{bmatrix}
        \bm{f}_{\rm{\lambda}}\\
        \bm{\tau}_{\rm{\lambda}}
    \end{bmatrix}
    =
    Q(\bm{q}) \bm{\lambda}_{\rm{full}},\\
    & && \lambda_{i,\rm{min}} \leq \lambda_{i} \leq \lambda_{i,\rm{max}}, \quad i = 1, \cdots, N.
\end{align}
We set $w_z = 1.0$ and $w_y = 10.0$, and define the weighting matrix $W$ as follows:
\begin{gather}
W =
\operatorname{diag}(
\underbrace{w_{\rm z},\,w_{\rm y},\,\dots,\,w_{\rm z},\,w_{\rm y}}_{2N_{\rm r}},
\underbrace{w_{\rm z},\,w_{\rm z},\,\dots,\,w_{\rm z}}_{N_{\rm s}}
).
\end{gather}
Here, the horizontal thrust components are penalized more heavily than the vertical ones, because horizontal thrust more easily excites vibrations of the flexible links.
However, when using the above thrust-minimization approach, the resulting thrust commands can oscillate, which often causes oscillations of the flexible links. To reduce such high-frequency variations, we propose an objective function that minimizes the difference from the thrust command at the previous control cycle, $\bm{\lambda}_{\rm{full,prev}}$:
\begin{align}
    \label{eq:lambda_opt_diff}
    & \underset{\bm{\lambda}_{\rm{full}}}{\text{minimize}} 
    & & 
        \begin{pmatrix} 
            \bm{\lambda}_{\rm{full}} - \bm{\lambda}_{\rm{full,prev}}
        \end{pmatrix}^{\rm{T}}
        W
        \begin{pmatrix} 
            \bm{\lambda}_{\rm{full}} - \bm{\lambda}_{\rm{full,prev}}
        \end{pmatrix}\\
    & \text{subject to} 
    && 
    \begin{bmatrix}
        \bm{f}_{\lambda}\\
        \bm{\tau}_{\lambda}
    \end{bmatrix}
    =
    Q(\bm{q}) \bm{\lambda}_{\rm{full}},\\
    & && \lambda_{i,\rm{min}} \leq \lambda_{i} \leq \lambda_{i,\rm{max}}, \quad i = 1, \cdots, N.
\end{align}

\subsection{Low-Rate Wrench-Allocation Matrix Updates}
\label{sec:reduce_update_hz_of_q}
As illustrated in \figref{control_system}, the desired linear and angular accelerations must be updated at a sufficiently high rate to ensure stable pose control. To suppress vibrations, however, we update the wrench-allocation matrix $Q$ at a lower rate, thereby filtering out high-frequency variations before thrust allocation. In our implementation, the desired accelerations and thrust commands are updated at over 200~Hz, while $Q$ is updated at below 20~Hz.

%% file: src/experiment.tex
\section{EXPERIMENT}

In this section, we first introduce prototypes of the proposed morphing aerial robot equipped with flexible continuum links. Next, using these prototypes, we evaluate the effectiveness of the vibration-suppressing control described in Sec.~\ref{control}. Finally, we demonstrate aerial manipulation tasks that leverage both shape adaptability and an expanded thrust wrench space, such as painting on curved surfaces, grasping objects with diverse shapes, and multi-object grasping. We further show that the robot's flexibility also contributes to impact absorption.

\subsection{Robot Platform}

\begin{figure}[t]
  \centering
  \includegraphics[width=\linewidth]{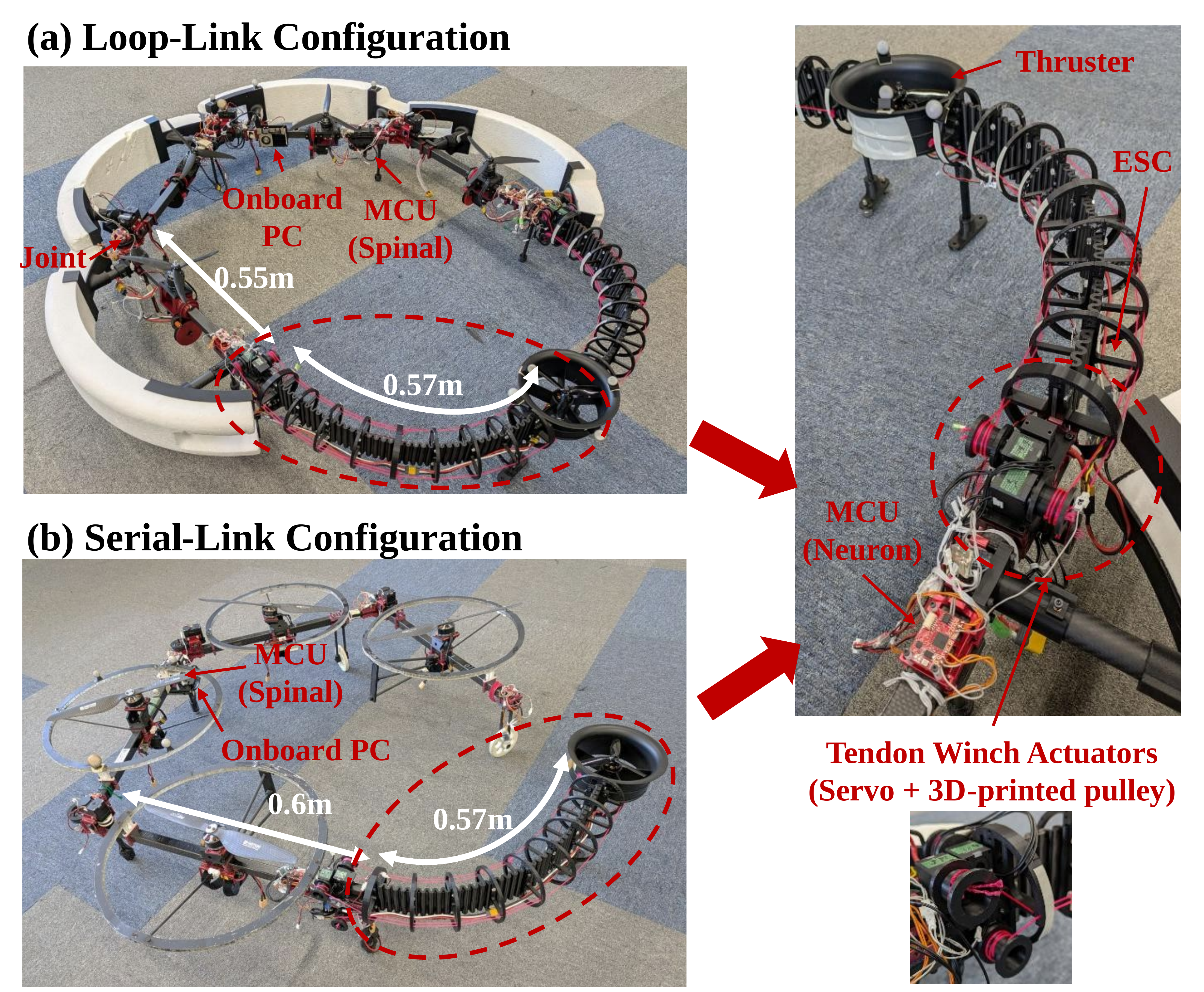}
  \caption{Prototypes of the proposed aerial robot}
  \label{figure:hardware_configuration}
  \vspace{-10pt}
\end{figure}
The overall hardware configuration is shown in \figref{hardware_configuration}. We developed two prototype variants: a loop-link configuration and a serial-link configuration. The loop-link configuration consists of four rigid links derived from Delta \cite{sugihara2024design} and two flexible continuum links, whereas the serial-link configuration consists of four rigid links derived from Hydrus \cite{zhao2017whole} and one flexible continuum link. The total mass is 4.6\,kg for the loop-link configuration and 4.5\,kg for the serial-link configuration. Both prototypes are equipped with a thruster on the flexible links.

The flexible links and their integrated thrusters are configured as follows. The thrusters consist of Cobra C-2217-12 KV1550 motors and 5-inch propellers, providing a maximum thrust of 20.2~N at 25.2~V. The flexible-link joints are tendon-driven and actuated by servo motors with 3D-printed pulleys; note that in the following manipulation tasks, the flexible links deform passively without actuation. The servos used are Dynamixel XC330-T288-T and Dynamixel XL430-W250-T.

We use a Khadas VIM4 onboard computer (Arm Cortex-A73 quad-core 2.2~GHz and Cortex-A53 quad-core 2.0~GHz) to construct the robot model from joint angles and solve the thrust-allocation problem using the OSQP quadratic programming solver \cite{osqp}. State estimation is performed using an IMU and a motion-capture system. Custom MCU boards, Spinal and Neuron \cite{anzai2017multilinked}, control the servos in each link and the motor ESCs.

\subsection{Evaluation of the Vibration-Suppressing Control}
\label{sec:experiment_vibration_suppression}

We evaluate the vibration-suppressing control introduced in Sec.~\ref{control} through flight experiments and demonstrate that both proposed methods are necessary for effective vibration suppression. The experiments were conducted for both the loop-link and serial-link configurations under vibration-prone conditions, such as in-air morphing and takeoff. To quantify the vibration of the flexible link, we focus on the thruster mounted on the flexible link, hereafter referred to as thrust5, and measure two signals associated with it: the thrust command \(\lambda_5\) and the roll angle \(\psi_{\mathrm{roll}}\), as shown in \figref{lambda_diff_opt}. We further analyze the frequency components of these data using the Fast Fourier Transform (FFT). In this evaluation, the position \(\bm{p}_5\) and rotor axis \(\bm{u}_5\) of rotor5 were obtained directly from the motion-capture system rather than estimated from joint angles.

\subsubsection{Thrust-Difference Minimization}
\begin{figure}[t]
  \centering
  \includegraphics[width=\columnwidth]{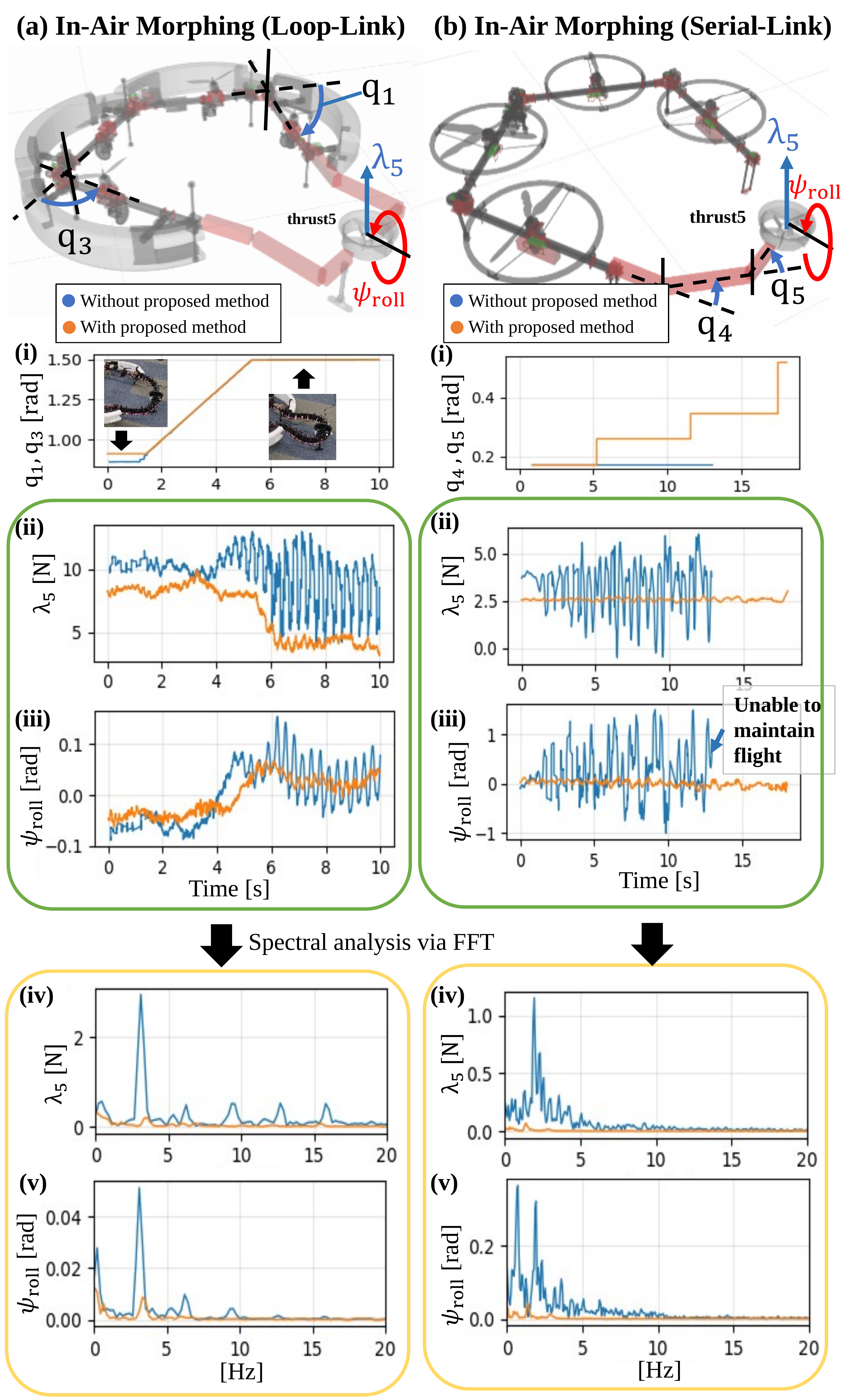}

  \caption{Experimental evaluation of the proposed thrust-difference minimization control (Sec.~\ref{sec:thrust_diff_min}). We compare the proposed objective, $\min (\bm{\lambda}-\bm{\lambda}_{\rm{prev}})^T W (\bm{\lambda}-\bm{\lambda}_{\rm{prev}})$ (shown in orange in plots), with the baseline thrust-minimization objective, $\min \bm{\lambda}^T W \bm{\lambda}$ (shown in blue in plots), for (a) in-air morphing with the loop-link configuration and (b) in-air morphing with the serial-link configuration. The proposed wrench-allocation strategy suppresses thrust oscillations and stabilizes thruster motion, enabling stable in-air morphing.}
  \label{figure:lambda_diff_opt}
  \vspace{-10pt}
\end{figure}

We first evaluate the effectiveness of thrust-difference minimization described in Sec.~\ref{sec:thrust_diff_min}. Throughout the following experiments, the second proposed control method, low-rate wrench-allocation matrix updates, was enabled.

For the loop-link configuration, an in-air morphing experiment was conducted, as shown in \figref{lambda_diff_opt}(a). During flight, the rigid-link joints $q_1$ and $q_3$ were actuated to 1.5~rad, causing large deformation of the flexible links and creating a vibration-prone configuration (\figref{lambda_diff_opt}(a)(i)). As indicated by the blue plots in \figref{lambda_diff_opt}(a)(ii)--(v), without thrust-difference minimization, both the thrust command $\lambda_5$ and the roll angle of thrust5 $\psi_{\mathrm{roll}}$ exhibited large oscillations, with dominant spectral peaks around 3~Hz. The peak amplitudes were 2.87~N for $\lambda_5$ and 0.051~rad for $\psi_{\mathrm{roll}}$. In contrast, with thrust-difference minimization (orange plots), these oscillations were markedly suppressed, and the corresponding peak amplitudes were reduced to 0.19~N and 0.0085~rad, respectively. These results demonstrate that the proposed method enables stable flight even in a structurally unstable configuration.

For the serial-link configuration, an in-air morphing experiment was conducted, as shown in \figref{lambda_diff_opt}(b). The flexible link was deformed by wire actuation to approximately $q_4 + q_5 = 1.0$~rad. As indicated by the blue plots in \figref{lambda_diff_opt}(b)(ii)--(v), without thrust-difference minimization, both the thrust command $\lambda_5$ and the roll angle of thrust5 $\psi_{\mathrm{roll}}$ exhibited large oscillations, with dominant spectral peaks around 2~Hz. The peak amplitudes were 1.154~N for $\lambda_5$ and 0.321~rad for $\psi_{\mathrm{roll}}$. These oscillations made continued flight difficult, and the robot had to land at around 13~s. In contrast, with thrust-difference minimization (orange plots), the peak amplitudes were reduced to 0.064~N for $\lambda_5$ and 0.042~rad for $\psi_{\mathrm{roll}}$, corresponding to a 87\% reduction in the $\psi_{\mathrm{roll}}$ amplitude. As a result, stable in-air morphing was achieved.

\subsubsection{Low-Rate Wrench-Allocation Matrix Updates}
\begin{figure}[t]
  \centering
  \includegraphics[width=\columnwidth]{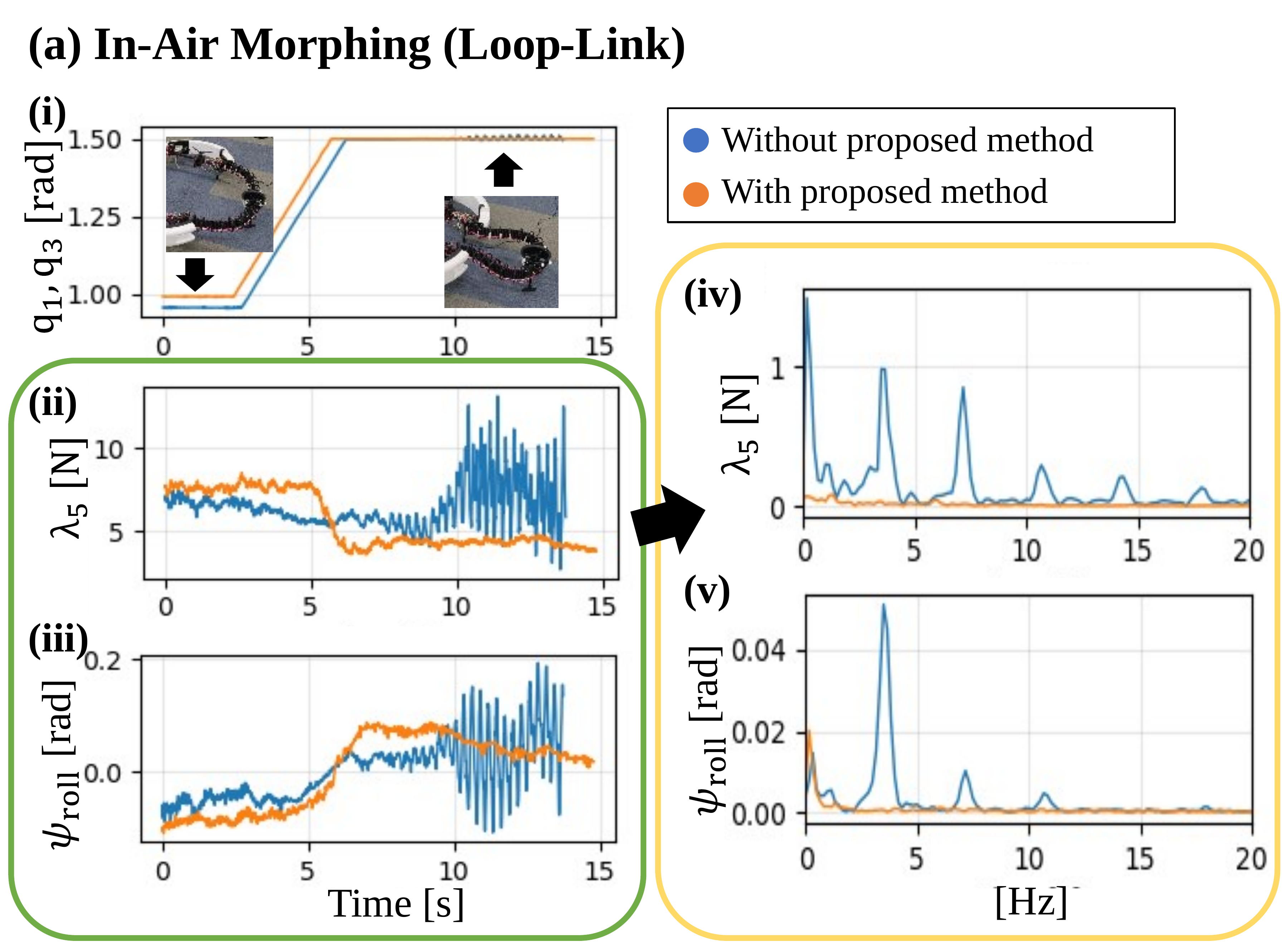}
  
  \caption{Experimental evaluation of the proposed control strategy that reduces the update frequency of the wrench-allocation matrix \(Q\) (Sec.~\ref{sec:reduce_update_hz_of_q}). The proposed strategy, in which \(Q\) is updated at 20~Hz or lower (orange plots), is compared with a baseline strategy in which \(Q\) is updated at every control cycle, i.e., at 200~Hz or higher (blue plots), under (a) in-air morphing in the loop-link configuration. The proposed strategy suppresses thrust oscillations and enables stable in-air morphing.}
  \vspace{-10pt}
  \label{figure:update_hz_of_q}
\end{figure}

We next evaluate the effectiveness of reducing the update frequency of the wrench-allocation matrix $Q$. Throughout the following experiments, the first proposed control method, thrust-difference minimization, was enabled.

For the loop-link configuration, an in-air morphing experiment was conducted, as shown in \figref{update_hz_of_q}(a). As indicated by the blue plots in \figref{update_hz_of_q}(a)(ii)--(v), when $Q$ was updated at 200~Hz, both $\lambda_5$ and $\psi_{\mathrm{roll}}$ exhibited large oscillations. Their dominant spectral peaks appeared around 3.5~Hz, with amplitudes of 0.985~N and 0.05~rad, respectively. In contrast, when the update rate of $Q$ was reduced (orange plots), these oscillations were greatly suppressed, and the spectral peaks around 3.5~Hz decreased to 0.031~N and 0.0006~rad, respectively. This enabled stable flight even in a largely deformed configuration.

For the serial-link configuration, a takeoff experiment was conducted. When $Q$ is updated at a high rate, large oscillations appeared immediately after takeoff, resulting in failed takeoff. In contrast, when the update frequency of $Q$ is reduced, the link still exhibited an initial transient oscillation after takeoff, but they were damped out over time. As a result, stable takeoff and subsequent flight were achieved. See the attached video for a visual comparison of the experiment.

\subsection{Aerial Painting on a Curved Surface}
\begin{figure}[t]
  \centering
  \includegraphics[width=\linewidth]{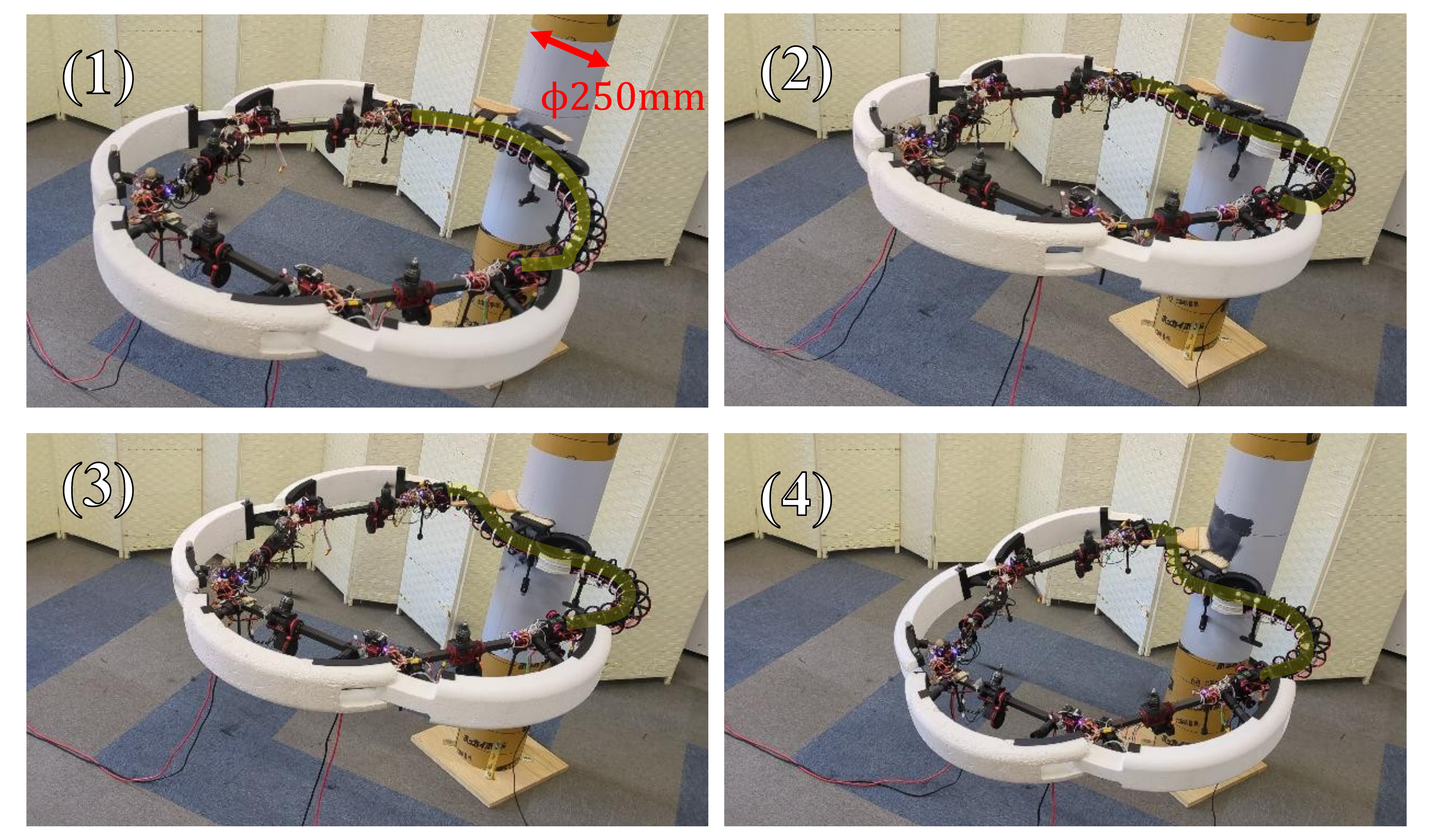}
  \caption{Demonstration of painting on a curved surface by utilizing the passive deformation of the flexible link. The flexible link passively deforms and conforms to a cylinder with a diameter of 250~mm, enabling stable flight while performing the painting task.}
  \label{figure:paint_demo}
  \vspace{-15pt}
\end{figure}

We conducted an aerial painting experiment on a curved surface, as shown in \figref{paint_demo}. For this task, the flexible link was equipped with brushes. Starting from an O-shaped configuration in flight, the flexible link was pressed against the target object, a cylinder with a diameter of 250~mm (\figref{paint_demo}(1)). The flexible link then passively deformed into an M-shaped configuration along the surface (\figref{paint_demo}(3)). In this experiment, the flexible links deformed by up to 0.2~m. Despite the large deformation, the robot maintained stable flight and successfully painted the surface. Aerial painting requires the end-effector to conform to the target geometry while generating sufficient moment to counteract contact forces. By leveraging both shape adaptability and an expanded wrench space, the robot successfully performed painting on a curved surface without requiring prior knowledge of the target geometry. However, brush--surface friction was not explicitly modeled, which led to attitude deviations. Incorporating this friction effect into the model remains an important direction for future work.

\subsection{Grasping Objects of Various Shapes}
\begin{figure}[t]
  \centering
  \includegraphics[width=\linewidth]{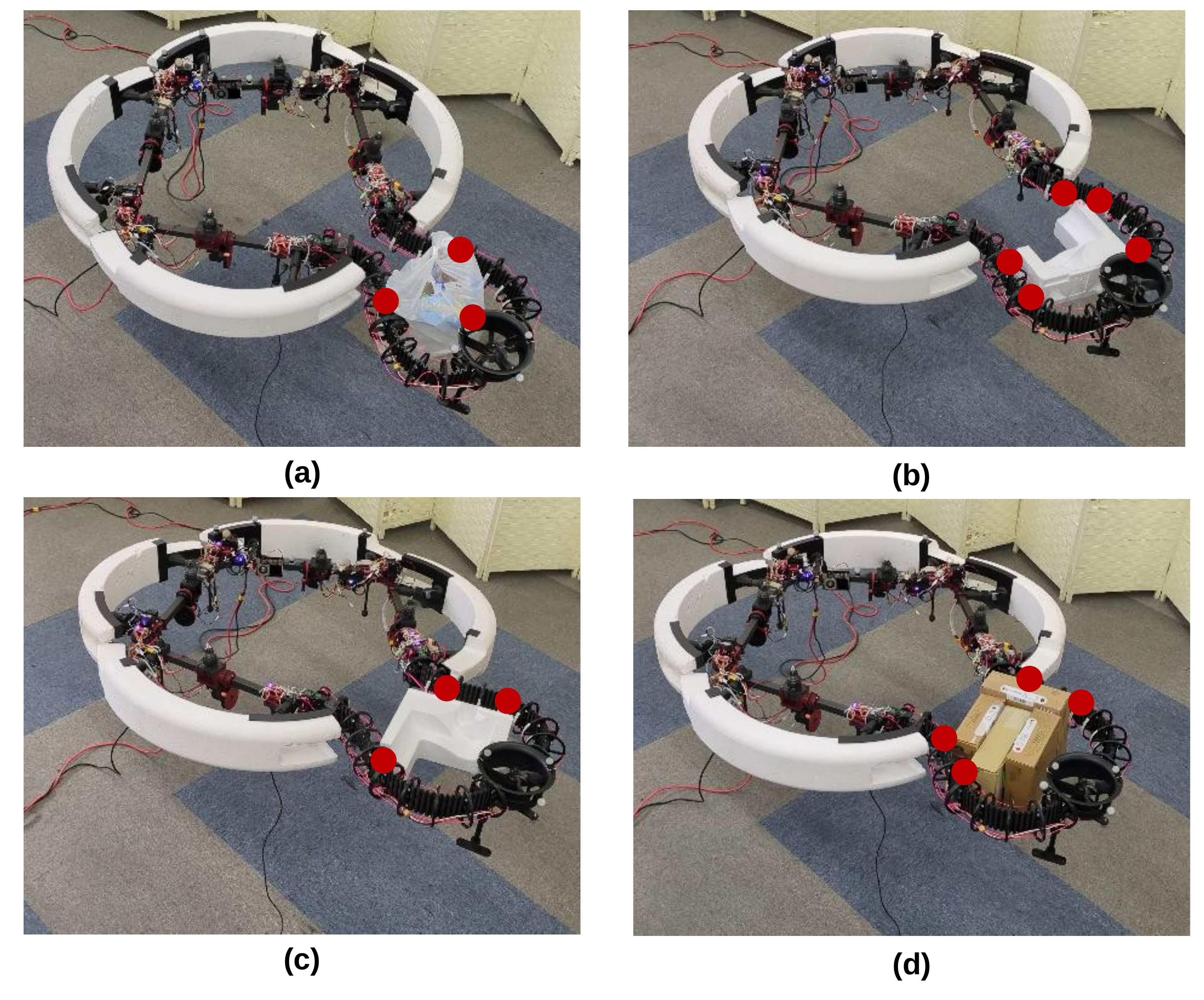}
  \caption{Demonstration of grasping objects of various shapes. (a) A plastic bag containing a randomly placed plastic bottle, (b)(c) L-shaped objects, and (d) a rectangular box were grasped. The red dots indicate the contact points with the grasped objects, showing that the gripper conforms to each object's geometry and establishes distributed contacts for stable grasping.}
  \label{figure:hold_various_objects}
\end{figure}

We conducted grasping experiments with objects of various shapes, as shown in \figref{hold_various_objects}. After achieving hovering, the rigid-link joints were actuated to close the flexible link around the objects. The flexible link passively conformed to the object geometry, enabling stable grasping. The mass of each target object was measured in advance and incorporated into the robot model. We successfully grasped a plastic bag containing a loosely placed plastic bottle (\figref{hold_various_objects}(a)), L-shaped objects (\figref{hold_various_objects}(b), (c)), and a rectangular box (\figref{hold_various_objects}(d)).

\begin{figure}[t]
  \centering
  \includegraphics[width=\linewidth]{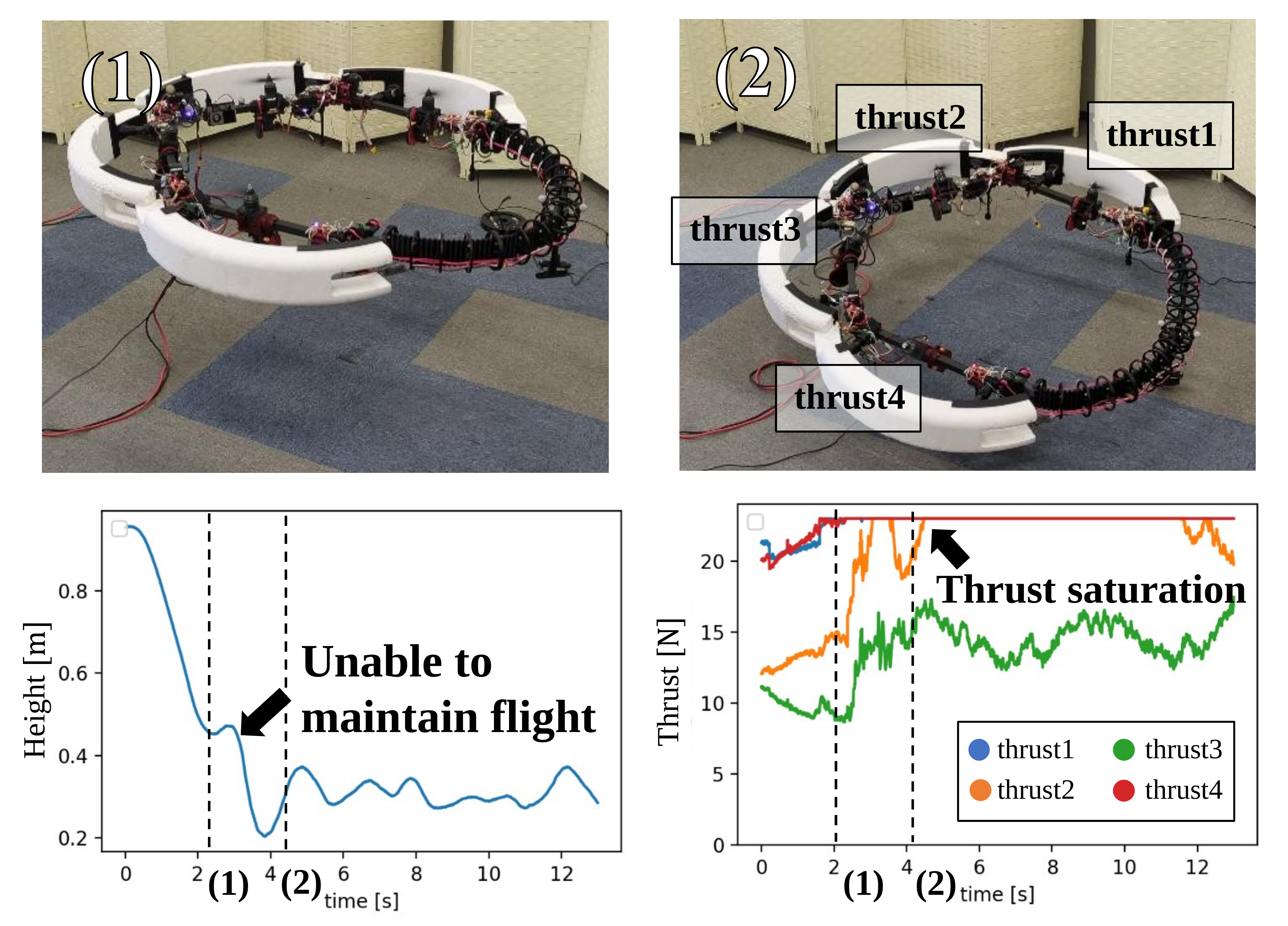}
  \caption{Flight without the thruster mounted on the flexible link. The robot could not generate sufficient thrust to maintain stable flight, particularly during upward acceleration, and eventually crashed.}
  \label{figure:remove_thrust5}
  \vspace{-10pt}
\end{figure}

Object grasping is a promising application of aerial robots \cite{Ruggiero2018aerialmanipulation}. However, robust grasping of diverse objects is difficult with rigid links because they lack the flexibility to conform to various object shapes. In contrast, soft manipulators can exploit structural compliance to passively adapt to different shapes, thereby enabling more robust grasping.

In addition, during aerial transport, the robot must generate sufficient thrust to compensate for the moments induced by the payload mass. In the loop-link configuration, without the thruster on the flexible link, the platform already operates near its thrust saturation even without carrying an additional payload, making stable flight difficult. In particular, the robot was unable to generate sufficient thrust during upward acceleration, which led to a crash, as shown in \figref{remove_thrust5}. Adding a thruster to the flexible links successfully addresses this limitation, and we achieved grasp-and-transport of payloads up to 683~g. We attribute the successful grasping of objects with diverse shapes to the combination of (i) the shape adaptability of the flexible link and (ii) sufficient thrust provided by the distributed thruster.

\subsection{Multi-Object Grasping}
\begin{figure}[t]
  \centering
  \includegraphics[width=\linewidth]{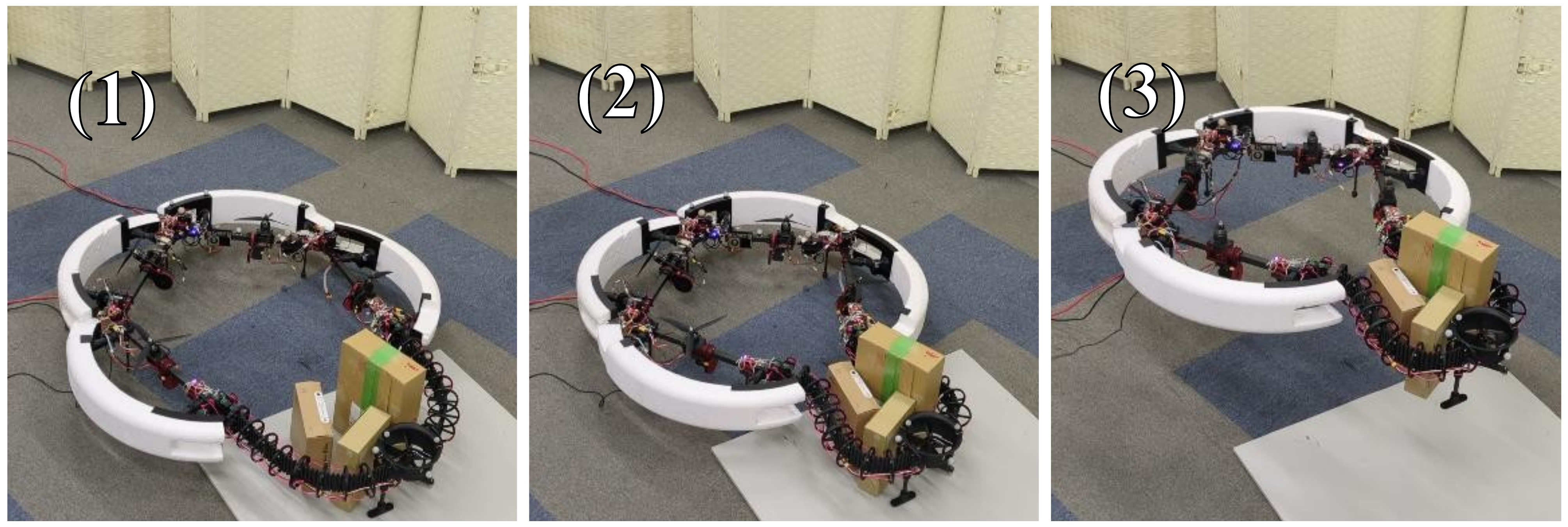}
  \caption{Demonstration of multi-object grasping using the flexible link. The flexible link passively deforms and closely adheres to multiple objects of various shapes, enabling stable multi-object grasping. By utilizing the thrust on the flexible link, the robot successfully took off while maintaining the grasp on the objects.}
  \label{figure:hold_and_takeoff}
  \vspace{-10pt}
\end{figure}

We demonstrated grasping of multiple objects placed in an unstructured arrangement, as shown in \figref{hold_and_takeoff}. Three boxes of different sizes were first surrounded by the flexible link. The rigid-link joints were actuated to close the flexible link and grasp all the boxes together. By passively conforming to the objects and applying distributed contact forces from multiple directions, the robot achieved stable multi-object grasping. With the thruster mounted on the flexible link, the robot successfully took off and performed aerial transport. The total mass of the grasped boxes was 388~g; this value was measured in advance and incorporated into the robot model. As discussed above, achieving stable flight under this condition is difficult without the thruster on the flexible link.

Multi-object grasping is generally challenging because of the high uncertainty in object shapes and poses. By leveraging both shape adaptability and an expanded wrench space, the proposed robot can grasp and transport multiple objects in a single grasp even in unstructured settings.

\subsection{Impact Absorption during Flight}
The flexible-link structure also contributes to impact absorption during flight. We compared the robot’s response when an impact was applied to a rigid link versus a flexible link. In the experiment, a 2.5~kg weight was released from a height of approximately 850~mm, swung as a pendulum, and collided with the robot while it was flying at a height of approximately 550~mm, corresponding to an impact speed of about 1.71~m/s. When the rigid link was struck, the joint driven by the pulley and timing belt skipped by 23.6~deg, resulting in mechanical damage. In contrast, when the flexible link was struck, the compliant structure absorbed the impact, and the robot maintained stable flight. See the attached video for a visual comparison of the experiment.

%% file: src/conclusion.tex
\section{CONCLUSIONS}

In this study, we proposed a morphing aerial robot with flexible continuum links that achieves both high environmental adaptability and an expanded achievable wrench set. The lightweight composite leaf-spring structure designed for the flexible continuum links suppresses sagging and vertical oscillations. In addition, we developed two vibration-suppressing control strategies: (i) thrust-difference minimization and (ii) low-rate updates of the wrench-allocation matrix. In flight experiments, these strategies further reduced flexible-link oscillations, achieving up to a 87\% reduction in the dominant spectral peak of the roll angle of the flexible-link thruster, and enabled successful takeoff and stable in-air morphing. Together, the proposed hardware design and control methods enabled stable flight and versatile aerial manipulation, thereby validating the effectiveness of flexible continuum links with integrated thrusters.

In future work, it will be important to increase the number of flexible continuum links and thrusters mounted on flexible section, so as to further enhance both shape adaptability and feasible wrench space. This is expected to enable more advanced manipulation tasks, such as painting on more complex geometries and grasping larger objects. However, longer flexible continuum links are more prone to oscillations, which can destabilize flight. Therefore, improved design and control strategies will be necessary to suppress these vibrations.

%% file: main.bib
@article{park2018odar,
  author={Park, Sangyul and others},
  journal={IEEE/ASME Transactions on Mechatronics}, 
  title={ODAR: Aerial Manipulation Platform Enabling Omnidirectional Wrench Generation}, 
  year={2018},
  volume={23},
  number={4},
  pages={1907-1918},
  doi={10.1109/TMECH.2018.2848255}}

@inproceedings{osqp,
  title={OSQP: An operator splitting solver for quadratic programs},
  author={Stellato, Bartolomeo and others},
  booktitle={2018 UKACC 12th international conference on control (CONTROL)},
  pages={339--339},
  year={2018},
  organization={IEEE}
}

@inproceedings{anzai2017multilinked,
  title={Multilinked multirotor with internal communication system for multiple objects transportation based on form optimization method},
  author={Anzai, Tomoki and others},
  booktitle={2017 IEEE/RSJ International Conference on Intelligent Robots and Systems (IROS)},
  pages={5977--5984},
  year={2017},
  organization={IEEE}
}

@article{tognon2019truly,
  title={A truly-redundant aerial manipulator system with application to push-and-slide inspection in industrial plants},
  author={Tognon, Marco and others},
  journal={IEEE Robotics and Automation Letters},
  volume={4},
  number={2},
  pages={1846--1851},
  year={2019},
  publisher={IEEE}
}

@article{Ruggiero2018aerialmanipulation,
  author={Ruggiero, Fabio and others},
  journal={IEEE Robotics and Automation Letters}, 
  title={Aerial Manipulation: A Literature Review}, 
  year={2018},
  volume={3},
  number={3},
  pages={1957-1964},
  doi={10.1109/LRA.2018.2808541}
}

@article{walker2013continuous,
  title={Continuous backbone “continuum” robot manipulators},
  author={Walker, Ian D},
  journal={International Scholarly Research Notices},
  volume={2013},
  number={1},
  pages={726506},
  year={2013},
  publisher={Wiley Online Library}
}

@article{shintake2018soft,
  title={Soft robotic grippers},
  author={Shintake, Jun and others},
  journal={Advanced materials},
  volume={30},
  number={29},
  pages={1707035},
  year={2018},
  publisher={Wiley Online Library}
}

@article{Ubellacker2024softdrone,
  title={High-speed aerial grasping using a soft drone with onboard perception},
  author={Ubellacker, Samuel and others},
  journal={npj Robotics},
  volume={2},
  number={1},
  pages={5},
  year={2024},
  publisher={Nature Publishing Group UK London}
}

@article{ZhichaoLiu2022softgripper,
  title={Safely catching aerial micro-robots in mid-air using an open-source aerial robot with soft gripper},
  author={Liu, Zhichao and others},
  journal={Frontiers in Robotics and AI},
  volume={9},
  pages={1030515},
  year={2022},
  publisher={Frontiers Media SA}
}

@article{wang2025spirobs,
  title={SpiRobs: Logarithmic spiral-shaped robots for versatile grasping across scales},
  author={Wang, Zhanchi and others},
  journal={Device},
  volume={3},
  number={4},
  year={2025},
  publisher={Elsevier}
}

@article{ruiz2022sophie,
  title={Sophie: Soft and flexible aerial vehicle for physical interaction with the environment},
  author={Ruiz, Fernando and others},
  journal={IEEE Robotics and Automation Letters},
  volume={7},
  number={4},
  pages={11086--11093},
  year={2022},
  publisher={IEEE}
}

@article{zhao2022forceful,
  title={Forceful valve manipulation with arbitrary direction by articulated aerial robot equipped with thrust vectoring apparatus},
  author={Zhao, Moju and others},
  journal={IEEE Robotics and Automation Letters},
  volume={7},
  number={2},
  pages={4893--4900},
  year={2022},
  publisher={IEEE}
}

@inproceedings{zhao2017whole,
  title={Whole-body aerial manipulation by transformable multirotor with two-dimensional multilinks},
  author={Zhao, Moju and others},
  booktitle={2017 IEEE international conference on robotics and automation (ICRA)},
  pages={5175--5182},
  year={2017},
  organization={IEEE}
}

@article{sugihara2024design,
  author={Sugihara, Kazuki and others},
  journal={IEEE Transactions on Robotics}, 
  title={Design, Control, and Motion Strategy for DELTA: Transformable Multilink Multirotor for Air-Ground Hybrid Locomotion and Manipulation (early access)}, 
  year={2026},
  volume={},
  number={},
  pages={},
  doi={10.1109/TRO.2026.3706558}}

@article{webster2010design,
  title={Design and kinematic modeling of constant curvature continuum robots: A review},
  author={Webster III, Robert J and others},
  journal={The International Journal of Robotics Research},
  volume={29},
  number={13},
  pages={1661--1683},
  year={2010},
  publisher={SAGE Publications Sage UK: London, England}
}

@article{fumagalli2014developing,
  title={Developing an aerial manipulator prototype: Physical interaction with the environment},
  author={Fumagalli, Matteo and others},
  journal={IEEE Robotics \& Automation Magazine},
  volume={21},
  number={3},
  pages={41--50},
  year={2014},
  publisher={IEEE}
}

@inproceedings{hunt20143d,
  title={3D printing with flying robots},
  author={Hunt, Graham and others},
  booktitle={2014 IEEE international conference on robotics and automation (ICRA)},
  pages={4493--4499},
  year={2014},
  organization={IEEE}
}

@inproceedings{papachristos2014efficient,
  title={Efficient force exertion for aerial robotic manipulation: Exploiting the thrust-vectoring authority of a tri-tiltrotor uav},
  author={Papachristos, Christos and others},
  booktitle={2014 IEEE international conference on robotics and automation (ICRA)},
  pages={4500--4505},
  year={2014},
  organization={IEEE}
}

@article{broers2022design,
  title={Design and testing of a bioinspired lightweight perching mechanism for flapping-wing MAVs using soft grippers},
  author={Broers, Krispin CV and others},
  journal={IEEE Robotics and Automation Letters},
  volume={7},
  number={3},
  pages={7526--7533},
  year={2022},
  publisher={IEEE}
}
